\documentclass{article}

\usepackage{microtype}
\usepackage{graphicx}
\usepackage{subfigure}
\usepackage{booktabs} 

\usepackage{hyperref}
\usepackage{enumitem}

\usepackage[table]{xcolor}
\usepackage{xspace}
\usepackage{multirow}
\usepackage{nicefrac}

\usepackage[accepted]{icml2025}

\usepackage{amsmath}
\usepackage{amssymb}
\usepackage{mathtools}
\usepackage{amsthm}

\usepackage{amsmath,amsfonts,bm}

\def\eqref#1{equation~\ref{#1}}

\def\1{\bm{1}}

\def\va{{\bm{a}}}

\def\vh{{\bm{h}}}

\def\vm{{\bm{m}}}

\def\vv{{\bm{v}}}

\def\vx{{\bm{x}}}

\DeclareMathAlphabet{\mathsfit}{\encodingdefault}{\sfdefault}{m}{sl}
\SetMathAlphabet{\mathsfit}{bold}{\encodingdefault}{\sfdefault}{bx}{n}

\newcommand{\softmax}{\mathrm{softmax}}

\newcommand{\ours}{{VISTA}\xspace}
\definecolor{lightgray}{gray}{0.9}

\usepackage[capitalize,noabbrev]{cleveref}

\theoremstyle{plain}

\theoremstyle{definition}

\theoremstyle{remark}

\usepackage[textsize=tiny]{todonotes}

\icmltitlerunning{VISTA: Visual Information Steering with Token-logit \textbf{A}ugmentation}

\begin{document}

\twocolumn[
\icmltitle{The Hidden Life of Tokens: Reducing Hallucination of Large Vision-Language Models via Visual Information Steering}



\icmlsetsymbol{equal}{*}




\begin{icmlauthorlist}
\icmlauthor{Zhuowei Li}{1}
\icmlauthor{Haizhou Shi}{1}
\icmlauthor{Yunhe Gao}{1,2}
\icmlauthor{Di Liu}{1}
\icmlauthor{Zhenting Wang}{1}
\icmlauthor{Yuxiao Chen}{1}
\icmlauthor{Ting Liu}{3}
\icmlauthor{Long Zhao}{3}
\icmlauthor{Hao Wang}{1}
\icmlauthor{Dimitris N. Metaxas}{1}
\end{icmlauthorlist}

\icmlaffiliation{1}{Rutgers University}
\icmlaffiliation{2}{Stanford University}
\icmlaffiliation{3}{Google DeepMind}

\icmlcorrespondingauthor{Zhuowei Li}{zhuowei.li@rutgers.edu}
\icmlcorrespondingauthor{Haizhou Shi}{haizhou.shi@rutgers.edu}

\icmlkeywords{Machine Learning, ICML}

\vskip 0.3in
]



\printAffiliationsAndNotice{}  

\begin{abstract}
Large Vision-Language Models (LVLMs) can reason effectively over both textual and visual inputs, but they tend to hallucinate syntactically coherent yet visually ungrounded contents. In this paper, we investigate the internal dynamics of hallucination by examining the tokens logits ranking throughout the generation process, revealing three key patterns in how LVLMs process information: 
\emph{(1) gradual visual information loss} -- visually grounded tokens gradually become less favored throughout generation, and 
\emph{(2) early excitation} -- semantically meaningful tokens achieve peak activation in the layers earlier than the final layer.
\emph{(3) hidden genuine information} -- visually grounded tokens though not being eventually decoded still retain relatively high rankings at inference.
Based on these insights, we propose \textbf{\ours} (\textbf{V}isual \textbf{I}nformation \textbf{S}teering with \textbf{T}oken-logit \textbf{A}ugmentation), a training-free inference-time intervention framework that reduces hallucination while promoting genuine information. \ours works by combining two complementary approaches: reinforcing visual information in activation space and leveraging early layer activations to promote semantically meaningful decoding. 
Compared to existing methods, \ours requires no external supervision and is applicable to various decoding strategies. 
Extensive experiments show that \ours on average reduces hallucination by about 40\% on evaluated open-ended generation task, and it consistently outperforms existing methods on four benchmarks across four architectures under three decoding strategies. Code is available at \url{https://github.com/LzVv123456/VISTA}. 
\end{abstract}

\section{Introduction}
\label{sec:intro}

Large Vision-Language Models (LVLMs)~\cite{instructblip, qwen, shikra, minigpt4, llava} have revolutionized multimodal AI by enabling seamless integration of visual and textual information, powering applications from interactive assistance to autonomous systems~\cite{ lin2023video, yang2024gpt4tools, lai2024lisa}. However, LVLMs frequently hallucinate semantically coherent yet visually ungrounded contents, hindering their reliability in real-world applications.

Though LVLM hallucination is considered multifaceted~\cite{liu2024survey}, a critical cause stems from the overwhelming influence of language priors over visual contexts, and has been studied from the perspective of attention patterns~\cite{opera, pai} and distribution divergence within logits space~\cite{vcd, favero2024multi}. Despite these insights, it remains unclear how hallucination emerges and propagates internally during the course of generation. 

\begin{figure}[t]
    \centering
    \includegraphics[width=0.49\linewidth]{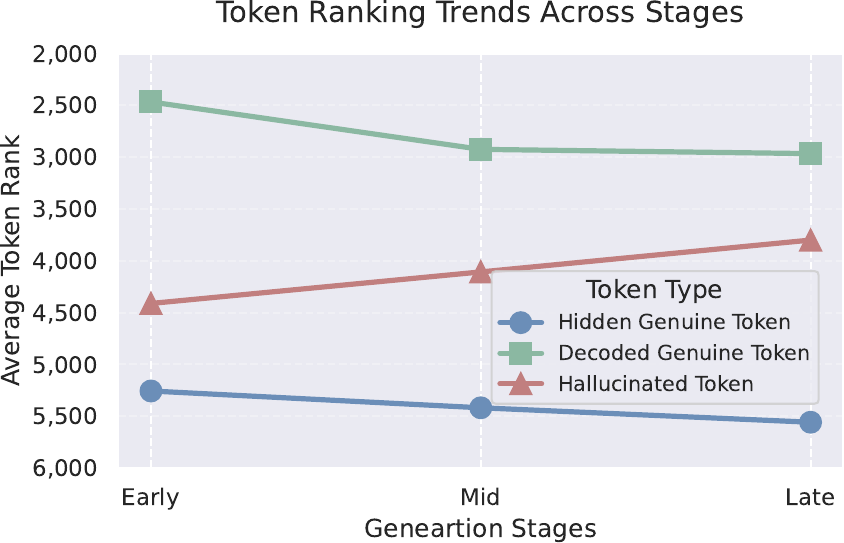}  %
    \hfill
    \includegraphics[width=0.49\linewidth]{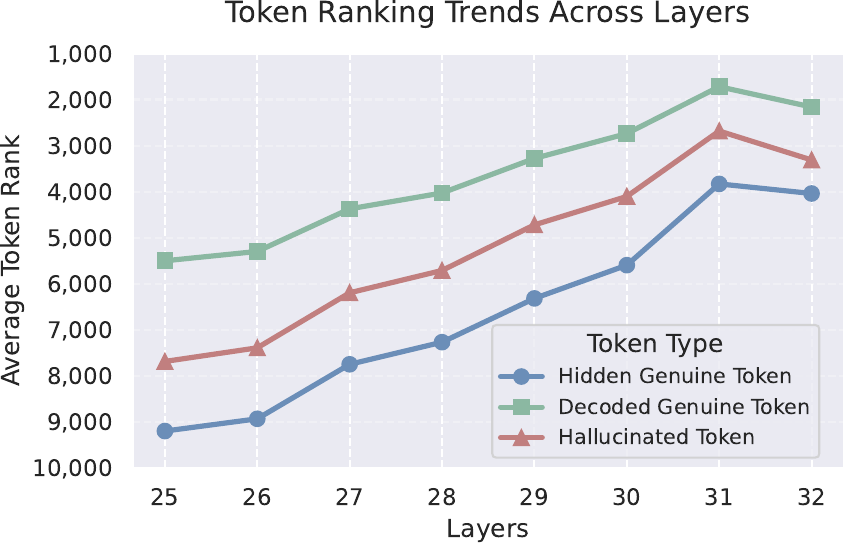}  
    \vspace{-10pt}
    \caption{Analysis of token logits ranking patterns across 500 randomly selected images from MSCOCO dataset. Higher ranking indicates higher generation probability. \textbf{Left:} Average token ranking from the last five layers, showing temporal progression across early, mid, and late generation stages. \textbf{Right:} Layer-wise evolution of token rankings averaged across all time steps, demonstrating early-excitation phenomenon.}
    \label{fig:global_view}
    \vspace{-20pt}
\end{figure}

\begin{figure}[t]
  \centering
   \includegraphics[width=1.0\linewidth]{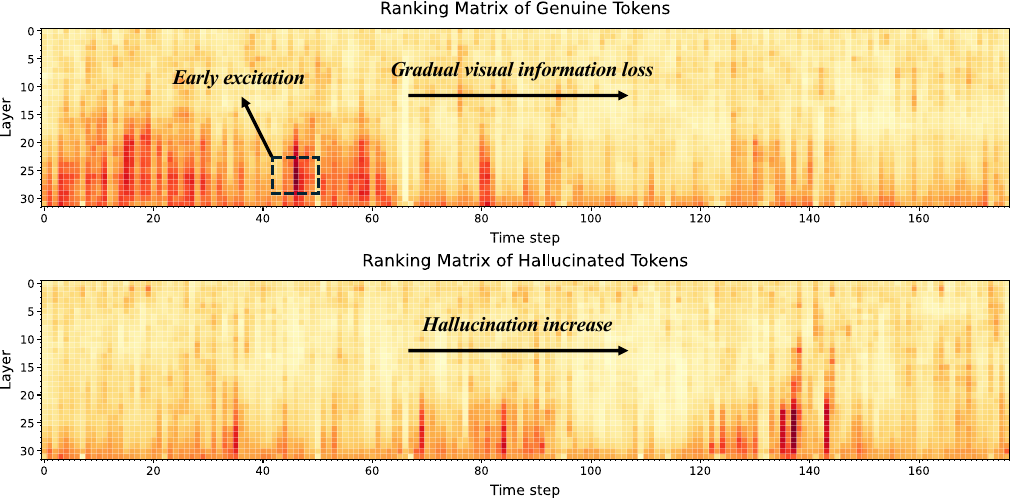}
   \vspace{-25pt}
   \caption{Token ranking heatmaps for a representative image, demonstrating the evolution of token rankings across model layers (vertical axis) and generation steps (horizontal axis). Darker colors indicate higher ranking. The visualization reveals both gradual visual information loss and early excitation phenomena.}
   \label{fig:rank_matrix}
   \vspace{-15pt}
\end{figure}

\textbf{Inspecting Token Dynamics in LVLMs.} 
In this work, we take a novel perspective by examining LVLM's generation dynamics through the lens of token logits ranking. Given an image and its corresponding description (produced by an LVLM), we identify three categories of tokens (elaborated in Sec.~\ref{sec:analysis}): 
\begin{itemize}[nosep,leftmargin=16pt]
    \item \emph{Hidden Genuine Tokens} -- tokens that are missing in generated contents yet clearly visible from visual input; 
    \item \emph{Decoded Genuine Tokens} -- tokens that appear in continuation with visual groundings;
    \item \emph{Hallucinated Tokens} -- tokens extracted from the hallucinated contents within generation.
\end{itemize}
We then track each token type's corresponding logits rankings throughout the generation across temporal (Fig.~\ref{fig:global_view} left) and layer sequences (Fig.~\ref{fig:global_view} right). Our analysis makes three prominent observations: 
\begin{itemize}[nosep,leftmargin=16pt]
    \item \textbf{(OBS-1) Gradual Visual Information Loss.} As generation progresses, genuine token rankings gradually decline while hallucinated tokens are surfaced (see Fig.~\ref{fig:global_view} left and Fig.~\ref{fig:rank_matrix}). This aligns with recent findings~\cite{yue2024less, favero2024multi}, exhibiting increasing hallucination in late generation phase. We hypothesize that this occurs as accumulated language priors in residual streams progressively dilute visual information, leading to syntactically coherent but visually ungrounded generation.
    \item \textbf{(OBS-2) Early Excitation of Semantically Meaningful Tokens.} Semantically meaningful tokens~\footnote{We refer to categorized tokens as semantically meaningful tokens since we primarily focus on identifying objects, attributes, and relations as detailed in Appendix~\ref{appendix:token_analysis}.} exhibit peak excitation in penultimate layer~(Fig.~\ref{fig:global_view} right) or within a window of layers preceding the final layer~(Fig.~\ref{fig:rank_matrix}). In contrast, the final layer prioritizes functional tokens like ``\texttt{this}'', ``\texttt{a}'', and other stopwords, suggesting that the model's decision process may overemphasize syntactic elements in its final stage.
    \item \textbf{(OBS-3) Hidden Genuine Information.} LVLMs may perceive more visual clues than they express. We observe that hidden genuine tokens, though not eventually decoded, consistently maintain relatively high rankings (around 5K in a 32K vocabulary) during the course of generation (see Fig.~\ref{fig:global_view}).
\end{itemize}

\textbf{Reducing Hallucination of LVLMs.} Inspired by above findings, we propose \textbf{\ours} (\textbf{V}isual \textbf{I}nformation \textbf{S}teering with \textbf{T}oken-logit \textbf{A}ugmentation), a simple and novel training-free framework that can be applied on top of various decoding methods to reduce hallucination while promoting genuine information.
\ours introduces two complementary modules: \textbf{\textit{Visual Steering Vector~(VSV)}} that counteracts gradual visual information loss by extracting and reinforcing visual cues in activation space, and \textbf{\textit{Self-Logits Augmentation~(SLA)}} which utilizes early excitation patterns to prioritize semantically meaningful tokens. VSV and SLA work synergistically and can effectively mitigate hallucination in LVLMs.

\textbf{Technical Contributions.} This study presents the first systematic investigation of token dynamics in LVLMs through the lens of token logits ranking, revealing novel insights into how visual information is processed and potentially lost during generation. Building on these insights, we propose \ours, an inference-time intervention framework that incorporates two complementary modules. The effectiveness of \ours is validated through comprehensive experiments across multiple architectures (e.g., LLaVA, Shikra, MiniGPT-4, InstructBLIP) and evaluation protocols (open-ended generation, visual question answering), demonstrating significant reduction in hallucination (up to around 40\% for open-ended generation). Our approach is notably efficient as it requires no additional training or model modifications, making it readily applicable to existing LVLM deployments.

\section{Methodology}
\label{sec:method}

In this section, we first establish the notions and knowledge foundation in Sec.~\ref{sec:method-pre}, followed by the elaboration of token ranking analysis in Sec.~\ref{sec:token_analysis}. We then present the proposed VSV and SLA methods in Sec.~\ref{sec:method-vsv} and Sec.~\ref{sec:method-sla}, respectively. 

\subsection{Preliminaries}
\label{sec:method-pre}
\textbf{Conditional Generation of LVLMs.}
Suppose that an LVLM consists of a vision encoder and a cross-modal interface that projects visual inputs into a sequence of visual tokens $\bm{X}_v$. Given an input image, the complete prompt tokens $\bm{X}_c$ are constructed by concatenating system message tokens $\bm{X}_s$ (can be empty), visual tokens $\bm{X}_v$, and query tokens $\bm{X}_q$: $\bm{X}_c = \texttt{concat}(\bm{X}_s, \bm{X}_v, \bm{X}_q)$. At each time step $t$, the model samples a new token $x_t$ according to the probability distribution conditioned on both the input context $\bm{X}_c$ and previously generated tokens $\bm{X}_{<t}=\{x_i\}_{i=1}^{t-1}$:
\begin{equation}
x_t \sim p(\vx_t| \bm{X}_c, \bm{X}_{<t}) = \softmax(\mathcal{H}(\vh_{t-1}^L)),
\end{equation}
where $\mathcal{H}$ denotes the model's head layer and $\vh_{t-1}^L$ indicates the hidden state from the last layer $L$ at time step $t-1$. Above formulation suggests that the dilution or insufficiency of visual information in $\vh_{t-1}^L$ can bias the generation towards hallucination. 

\textbf{Residual Stream.} Taking the mathematical interpretation from \citet{elhage2021mathematical}, we view layer-wise hidden states as residual streams that evolve recursively:
\begin{equation}
\vh_t^l = \vh^{l-1}_t + \va^l_t + \vm^l_t,
\end{equation}
where $l$ is the layer index, $\va^l_t$ and $\vm^l_t$ represent the output activation of integrated multi-head attention (MHA) layer and feed-forward network (FFN), respectively. Within this framework, MHAs facilitate information fusion across different residual streams, while FFNs access and integrate learned parametric knowledge~\cite{geva-etal-2021-transformer, dai-etal-2022-knowledge}. Residual stream provides a natural interface for monitoring and controlling information flow, making it particularly suitable for hallucination analysis and mitigation.

\textbf{Logit Lens.} The head layer $\mathcal{H}$ is by default applied on top of last layer hidden states $\vh^L$. However, thanks to the gradual evolvement of hidden states within residual streams~\cite{dola}, applying $\mathcal{H}$ to hidden states of earlier layers $l<L$ remains effective, even without additional training~\cite{logitlen}. This practice is commonly referred to as ``logit lens'' and can be used to decipher intermediate states.

\subsection{Token Ranking Analysis}\label{sec:token_analysis}
To systematically investigate how visual information is processed during generation, we propose a token ranking analysis framework that tracks the relative importance of different tokens throughout the generation process.

\textbf{Identification of Target Tokens.} 
For each given image-description pair where the text description is generated by an LVLM (e.g., LLAVA-1.5~\cite{llava15}), we utilize gpt-4o~\cite{hurst2024gpt} as the oracle model to identify three categories of words referencing both visual and textual contents. A word is a 
\begin{itemize}[nosep,leftmargin=16pt]
    \item \emph{decoded genuine word} if it appears in the continuation and align with visual evidence;
    \item \emph{hidden genuine word} if it is visually evident but not included in the continuation;
    \item \emph{hallucinated word} if it appears in continuation but lacks visual grounding.
\end{itemize}   
Collected words are then tokenized to form our analysis sets. Implementation details are included in Appendix~\ref{appendix:token_analysis}.

\textbf{Token Ranking via Logit Lens.} 
To analyze token dynamics during generation, we apply the logit lens $\mathcal{H}(\vh^l_t)$ to each layer $l$ and time step $t$. Given token $x$, we calculate its ranking position among all possible tokens according to:
\begin{equation}
    R_{t}^l(x) = \text{rank}(\mathcal{H}(\vh^l_t), x),
\end{equation}
where $R_{t}^l(x)$ represents the position of token $x$ in the probability-ordered sequence of all tokens at time step $t$ and layer $l$. A lower rank indicates higher probability.
This operation produces a 2D ranking matrix for each token with vertical and horizontal axes indicating layers and time steps, respectively. We aggregate these matrices across tokens within each category to obtain category-specific ranking patterns, as shown in Fig.~\ref{fig:rank_matrix}.

For temporal analysis, we quantize time sequence into three equal-sized buckets, i.e., early, mid, and late, and compute average rankings within each bucket across 500 randomly sampled images from MS COCO dataset~\cite{lin2014microsoft}. This temporal view (Fig.~\ref{fig:global_view} left) reveals that visually grounded information is gradually sinking while hallucinated contents are surfaced. We further average rankings across all time steps within a layer to provide a layer-wise perspective (Fig.~\ref{fig:global_view} right), which exhibits that semantically meaningful tokens achieve peak excitation in the penultimate layer. We refer readers to Appendix~\ref{appendix:token_analysis_discussion} for a discussion of why token ranking analysis is desirable and its limitations.

\subsection{Visual Steering Vector (VSV)}
\label{sec:method-vsv}
Being aware of the challenge from gradual visual information loss, it is of critical importance to retain visual cues throughout the generation. 
A promising method involves increasing attention weights distributed on visual tokens~\cite{pai}. Nevertheless, this operation simultaneously introduces undesired parametric priors cumulated in residual streams of visual tokens.
Drawing inspiration from steering vectors in LLMs~\cite{turner2023activation, zou2023representation, liu2024incontext, li2024implicit}, we propose Visual Steering Vector~(VSV) to steer the generation of LVLM towards the direction with visual groundings without amplifying inherent language biases.

\begin{figure}[t]
  \centering
   \includegraphics[width=1.0\linewidth]{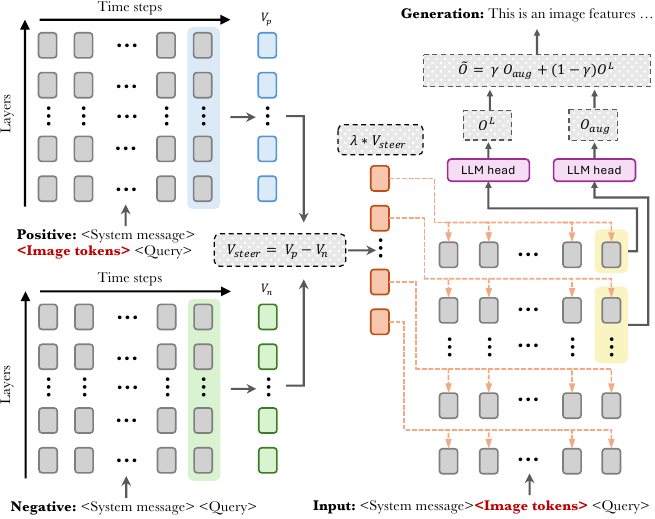}
   \vspace{-20pt}
   \caption{Architectural overview of \ours. \ours introduces two complementary mechanisms: VSV extracts and reinforces visual grounding information ($V_s$) at inference, and SLA leverages early-layer semantic information to guide token generation. \textit{Note: While three separate forward passes are shown for illustration purpose, they can be avoided in implementation.}}
   \label{fig:overview}
   \vspace{-10pt}
\end{figure}

\textbf{VSV Construction.} The core logic behind VSV is to extract a directional vector within activation space without introducing disturbing language priors. To this end, we construct VSV via a contrastive process using paired context sequences: a ``positive'' context $\bm{X}_p = \texttt{concat}(\bm{X}_s, \bm{X}_v, \bm{X}_q)$ containing visual tokens $\bm{X}_v$, and a ``negative'' counterpart $\bm{X}_n = \texttt{concat}(\bm{X}_s, \bm{X}_q)$ that discards visual tokens while preserving other elements. Both sequences are processed by a vectorization function $\mathcal{F}$, which forwards the given token sequence through the LVLM and takes the residual stream from the last token. Visual steering vector (VSV) can be computed as:
\begin{equation}
\bm{V}_{\text{steer}} = \bm{V}_p - \bm{V}_n = \{\vv_{\text{steer}}^l\}_{l=1}^L,
\end{equation}
where $\bm{V}_p = \mathcal{F}(\bm{X}_p)$ and $\bm{V}_n =  \mathcal{F}(\bm{X}_n)$. Here, $\vv_{\text{steer}}^l$ refers to the steering vector for layer $l$.

\textbf{Inference-time Intervention.} During inference, we inject the visual steering vector into the residual stream at each generation step:
\begin{equation}
    \Tilde{\vh}_t^l = \vh_t^l + \lambda \vv_s^l, \quad l\in[1, L],
\end{equation}
with $\lambda$ controlling the intervention strength, balancing visual fidelity against natural generation. To maintain stability, we normalize the modified hidden states:
\begin{equation}
\Tilde{\vh}_t^l = \Tilde{\vh}_t^l \cdot \frac{\|\vh_t^l\|_2}{\|\Tilde{\vh}_t^l\|_2}, \quad l\in[1, L].
\end{equation}

Unlike steering vectors in LLMs that capture abstract concept in an amortized fashion leveraging a group of contrastive pairs~\cite{zou2023representation}, VSV computes the steering vector per image basis to preserve crucial visual details unique to each input image.

\textbf{Remark on Difference with Existing Contrastive Strategy.}
Existing contrastive strategy \cite{li2022contrastive, vcd} necessitates a second negative logits distribution to contrast the final layer logits at all generation steps. VSV, on the other hand, extracts a steering vector priori using only the context tokens, and is applied to all layers (Fig.~\ref{fig:overview}). In practice, formatting prompt and general negative prompt (e.g., \texttt{Describe the image in detail.}) can be forwarded only once and cached for future usage, making VSV remarkably efficient (see Table~\ref{tab:latency}).

\subsection{Self-Logits Augmentation (SLA)}
\label{sec:method-sla}
Motivated by early excitation phenomenon (Fig.~\ref{fig:global_view} right), where semantically meaningful tokens show stronger activation in penultimate layer, we propose Self-Logits Augmentation (SLA) to promote the decoding of such tokens.

\textbf{Augmentation Logits.}
To elicit the rich semantic information present in the late layers, we calculate ``augmentation logits'' $\bm{o}_t^{\text{aug}}$ at each decoding time step $t$.
These are obtained by applying the head layer $\mathcal{H}$ to the hidden states of the $w$ layers prior to the final layer and averaging their logits:
\begin{equation} 
\bm{o}_t^{\text{aug}} = \frac{1}{w} \sum_{l = L - w}^{L-1} \mathcal{H}(\vh_t^l),
\end{equation}
where $w$ indicates the window size. In practice, larger $w$ leads to improved performance (see Table~\ref{tab:ab_window_size}). We hypothesize that this is due to the ever-evolving logits distributions within late layers, especially for those semantically meaningful tokens~\cite{dola}, and applying a larger $w$ smooths the distribution and provides nuanced information. 

\textbf{Logits Ensemble.} 
Augmentation logits are then integrated with the final layer logits through weighted aggregation:
\begin{equation} 
\tilde{\bm{o}}_t = (1-\gamma) \cdot \bm{o}_t^L + \gamma \cdot \bm{o}_t^{\text{aug}},
\end{equation}
where $\bm{o}_t^L = \mathcal{H}(\vh_t^L)$ is the logits of the last layer, and $\gamma \in [0, 1]$ is a constant coefficient controlling the influence of early-layer information--when $\gamma = 0$, the model reduces to standard generation, while $\gamma = 1$ would rely entirely on early-layer logits. The next token is then sampled from the output distribution $x_{t} \sim \softmax(\tilde{\bm{o}}_{t-1})$. Notably, SLA seeks to promote the decoding of more semantically meaningful tokens than truthful tokens as done in DoLa~\cite{dola}. It works synergistically with the VSV module, enhancing overall performance (see Fig.~\ref{fig:ab_shikra}).

\section{Experiments}\label{sec:experiment}
In this section, we empirically validate \ours across four architectures, three decoding strategies, and four benchmarks. We first present the experimental configuration~(Sec.~\ref{sec:experiment-setup}), followed by an extensive evaluation on hallucination-specific and general-purpose benchmarks~(Sec.~\ref{sec:experiment-main-obj} and \ref{sec:experiment-main-comp}). We then analyze VISTA's effectiveness in addressing the observed phenomena~(Sec.~\ref{sec:analysis}) and conclude with comprehensive ablation studies~(Sec.~\ref{sec:ablation}).

\begin{table*}[t]
    \centering
    \caption{CHAIR hallucination evaluation results. We compare \ours to state-of-the-art training-free methods that do not rely on external supervision. Maximum new token is set to 512. \textbf{Best} and \underline{second best} results are bolded and underlined, respectively. ``-'' indicates the result is not supported by released implementation.
    }
    \resizebox{0.95\textwidth}{!}{
    \begin{tabular}{llcccccccc}
        \toprule
        \multirow{2}{*}{\centering Decoding} & \multirow{2}{*}{\centering Method} & \multicolumn{2}{c}{LLAVA-1.5~\cite{llava15}} & \multicolumn{2}{c}{MiniGPT-4~\cite{minigpt4}} & \multicolumn{2}{c}{Shikra~\cite{shikra}} & \multicolumn{2}{c}{InstructBLIP~\cite{instructblip}} \\
        \cmidrule(lr){3-4} \cmidrule(lr){5-6} \cmidrule(lr){7-8} \cmidrule(lr){9-10}
        & & CHAIR$_\text{S}\downarrow$ & CHAIR$_\text{I}\downarrow$ & CHAIR$_\text{S}\downarrow$ & CHAIR$_\text{I}\downarrow$ & CHAIR$_\text{S}\downarrow$ & CHAIR$_\text{I}\downarrow$ & CHAIR$_\text{S}\downarrow$ & CHAIR$_\text{I}\downarrow$ \\
        \midrule
        \multirow{5}{*}{Greedy} 
        & Vanilla & 46.4 & 12.1 & 35.2 & 10.7 & 56.8 & 14.8 & \underline{38.0} & \underline{10.7} \\
        & DoLa ~\cite{dola} & 45.4 & 11.9 & - & - & 60.0 & 15.1 & - & - \\
        & VCD~\cite{vcd} & 47.4 & 13.0 & - & - & - & - & 45.8 & 12.8 \\
        & PAI~\cite{pai} & \underline{22.8} & \underline{7.0} & \underline{29.2} & \underline{10.9} & \underline{40.8} & \underline{11.0} & - & - \\
        & \cellcolor{lightgray}\textbf{\ours} (ours) & \cellcolor{lightgray}\bf 20.4 & \cellcolor{lightgray}\bf 6.9 & \cellcolor{lightgray}\bf 19.8 & \cellcolor{lightgray}\bf 6.0 & \cellcolor{lightgray}\bf 31.4 & \cellcolor{lightgray}\bf 9.7 & \cellcolor{lightgray}\bf 27.4 & \cellcolor{lightgray}\bf 8.1 \\
        \midrule
        \multirow{5}{*}{Beam Search} 
        & Vanilla & 49.0 & 12.5 & 33.0 & 11.0 & 53.8 & 14.4 & \underline{37.8} & \underline{10.7} \\
        & VCD~\cite{vcd} & 49.8 & 12.4 & - & - & - & - & 49.2 & 13.7 \\
        & OPERA~\cite{opera} & 45.2 & 12.4 & \underline{26.8} & \underline{9.3} & \underline{39.6} & 12.2 & 50.2 & 13.9 \\
        & PAI~\cite{pai} & \underline{22.3} & \underline{6.8} & 31.6 & 11.2 & 41.6 & \underline{10.4} & - & - \\
        & \cellcolor{lightgray}\textbf{\ours} (ours) & \cellcolor{lightgray}\bf 17.4 & \cellcolor{lightgray}\bf 6.3 & \cellcolor{lightgray}\bf 18.4 & \cellcolor{lightgray}\bf 6.4 & \cellcolor{lightgray}\bf 32.2 & \cellcolor{lightgray}\bf 9.5 & \cellcolor{lightgray}\bf 26.8 & \cellcolor{lightgray}\bf 7.8 \\
        \midrule
        \multirow{5}{*}{Nucleus Sampling} 
        & Vanilla & 53.2 & 15.1 & 34.8 & 11.2 & 56.4 & 15.9 & \underline{46.6} & \underline{13.1} \\
        & DoLa ~\cite{dola} & 47.2 & 14.0 & - & - & 56.6 & 16.3 & - & - \\
        & VCD~\cite{vcd} & 60.8 & 16.2 & - & - & - & - & 57.0 & 16.0 \\
        & PAI~\cite{pai} & \underline{30.2} & \underline{10.3} & \underline{31.8} & \underline{13.2} & \underline{43.2} & \underline{12.0} & - & - \\
        & \cellcolor{lightgray}\textbf{\ours} (ours) & \cellcolor{lightgray}\bf 24.0 & \cellcolor{lightgray}\bf 8.2 & \cellcolor{lightgray}\bf 18.4 & \cellcolor{lightgray}\bf 6.4 & \cellcolor{lightgray}\bf 31.8 & \cellcolor{lightgray}\bf 9.7 & \cellcolor{lightgray}\bf 29.4 & \cellcolor{lightgray}\bf 9.1 \\
        \bottomrule
    \end{tabular}
    }
    \label{tab:chair}
    \vspace{-10pt}
\end{table*}

\subsection{Experimental Setup}
\label{sec:experiment-setup}
\textbf{Model Architectures.} We evaluate VISTA on four representative LVLMs with distinct architectural designs: \textbf{LLAVA-1.5}~\cite{llava15} and \textbf{Shikra}~\cite{shikra}, which employ linear projections for visual-textual alignment, and \textbf{MiniGPT-4}~\cite{minigpt4} and \textbf{InstructBLIP}~\cite{instructblip}, which utilize Q-former~\cite{li2023blip} for cross-modal interaction.

\textbf{Decoding Strategies.} To demonstrate VISTA's versatility as an inference-time intervention method, we verify it across three widely used decoding protocols: (1) \textbf{greedy decoding}, which selects the highest probability token at each step, (2) \textbf{beam search} with a beam size of 5, maintaining multiple generation hypotheses, and (3) \textbf{nucleus sampling} with top-p=0.9. Temperature is fixed at 1.0 for all scenarios.

\textbf{Baselines.} Besides three vanilla decoding strategies, we compare VISTA with several SoTA hallucination mitigation methods that can operate without external supervision. \textbf{DoLa}~\cite{dola} signifies a internal (across layers) contrastive strategy within logits space, while \textbf{VCD}~\cite{vcd} presents a parallel contrastive method across time steps. \textbf{OPERA}~\cite{opera} is a powerful technique tailored for the beam search. \textbf{PAI}~\cite{pai} is another inference-time intervention method that is most comparable to ours. We reproduce all baseline results using identical evaluation data and settings (e.g., prompt, temperature). \textit{{Methods without official supports for certain architectures and decoding strategies are omitted to prevent implementation bias.}}

\textbf{Implementation Details.} We employ the following configuration across all experiments unless stated otherwise. The VSV strength parameter $\lambda$ is set to 0.17 for both LLAVA-1.5 and InstructBLIP, 0.1 for MiniGPT-4, and 0.12 for Shikra. The SLA mixing coefficient $\gamma$ is consistently set to 0.3, with a window size $w=5$ for aggregating early-layer logits. We search the hyperparameters of $\lambda$ and $\gamma$ on a holdout validation set containing 100 images from MSCOCO to balance between generation quality and hallucination reduction.

\begin{table*}[t]
    \centering
    \caption{Evaluation results on POPE benchmark across four LVLMs. Results show averaged accuracy and F1 scores computed across random, popular, and adversarial object splits. \textbf{Best} and \underline{second best} results are bolded and underlined, respectively.}
    \resizebox{0.95\textwidth}{!}{
    \begin{tabular}{llcccccccc}
        \toprule
        \multirow{2}{*}{\centering Decoding} & \multirow{2}{*}{\centering Method} & \multicolumn{2}{c}{LLAVA-1.5~\cite{llava15}} & \multicolumn{2}{c}{MiniGPT-4~\cite{minigpt4}} & \multicolumn{2}{c}{Shikra~\cite{shikra}} & \multicolumn{2}{c}{InstructBLIP~\cite{instructblip}} \\
        \cmidrule(lr){3-4} \cmidrule(lr){5-6} \cmidrule(lr){7-8} \cmidrule(lr){9-10}
        & & Avg. Accuracy $\uparrow$ & Avg. F1 $\uparrow$ & Avg. Accuracy $\uparrow$ & Avg. F1 $\uparrow$ & Avg. Accuracy $\uparrow$ & Avg. F1 $\uparrow$ & Avg. Accuracy $\uparrow$ & Avg. F1 $\uparrow$ \\
        \midrule
        \multirow{5}{*}{Greedy} 
        & Vanilla & 84.79 & 85.61 & \underline{76.76} & 76.82 & 81.32 & \underline{82.01} & 84.36 & 84.64 \\
        & DoLa~\cite{dola} & 84.92 & 85.67 & - & - & 81.13 & 81.94 & - & - \\
        & VCD~\cite{vcd} & 84.80 & 85.65  & - & - & - & - & \underline{84.81} & \underline{85.28} \\
        & PAI~\cite{pai} & \underline{85.85} & \underline{86.08} & 75.64 & \underline{77.57} & \underline{81.30} & 80.81 & - & - \\
        & \cellcolor{lightgray}\textbf{\ours} (ours) & \cellcolor{lightgray}\bf 86.15 & \cellcolor{lightgray}\bf 86.29 & \cellcolor{lightgray}\bf 77.06 & \cellcolor{lightgray}\bf 77.80 & \cellcolor{lightgray}\bf 82.44 & \cellcolor{lightgray}\bf 82.47 & \cellcolor{lightgray}\bf 84.87 & \cellcolor{lightgray}\bf 84.95 \\
        \midrule
        \multirow{5}{*}{Beam Search} 
        & Vanilla & 85.45 & 84.93 & 73.68 & 72.40 & 81.73 & 82.10 & 84.38 & 83.71 \\
        & VCD~\cite{vcd} & \underline{85.85} & 85.90 & - & - & - & - & 84.90 & 84.43 \\
        & OPERA~\cite{opera} & 85.68 & 85.83 & \underline{74.81} & \underline{75.42} & \underline{82.18} & \underline{82.49} & \underline{85.31} & \underline{85.51} \\
        & PAI~\cite{pai} & \bf 86.27 & \underline{85.91} & 73.83 & 74.63 & 81.90 & 81.08 & - & - \\
        & \cellcolor{lightgray}\textbf{\ours} (ours) & \cellcolor{lightgray}85.83 & \cellcolor{lightgray}\bf 85.95 &  \cellcolor{lightgray}\bf75.96 & \cellcolor{lightgray}\bf 77.17 & \cellcolor{lightgray}\bf 82.54 & \cellcolor{lightgray}\bf 82.52 & \cellcolor{lightgray}\bf 85.78 & \cellcolor{lightgray}\bf 85.74 \\
        \midrule
        \multirow{5}{*}{Nucleus Sampling} 
        & Vanilla & 81.26 & 82.40 & 60.56 & 62.04 & 78.94 & 80.18 & 78.83 & 79.74 \\
        & DoLa~\cite{dola} & 81.20 & 82.44 & - & - & \underline{79.49} & \underline{80.72} & - & - \\
        & VCD~\cite{vcd} & 81.08 & 82.22 & - & - & - & - & \underline{79.61} & \underline{80.43} \\
        & PAI~\cite{pai} & \underline{81.92} & \underline{83.16} & \underline{61.26} & \underline{63.40} & 79.25 & 79.87 & - & - \\
        & \cellcolor{lightgray}\textbf{\ours} (ours) & \cellcolor{lightgray}\bf 85.35 & \cellcolor{lightgray}\bf 85.54 & \cellcolor{lightgray}\bf 66.96 & \cellcolor{lightgray}\bf 68.05 & \cellcolor{lightgray}\bf 81.01 & \cellcolor{lightgray}\bf 81.15 & \cellcolor{lightgray}\bf 83.11 & \cellcolor{lightgray}\bf 83.27 \\
        \bottomrule
    \end{tabular}
    }
    \label{tab:pope}
    \vspace{-10pt}
\end{table*}

\subsection{Results on Object Hallucination Benchmarks}
\label{sec:experiment-main-obj}
We first evaluate \ours on two widely adopted benchmarks that assess object hallucination: CHAIR~\cite{rohrbach2018object} for open-ended generation and POPE~\cite{rohrbach2018object} for targeted visual question answering.

\textbf{CHAIR Evaluation.} 
Caption Hallucination Assessment with Image Relevance (CHAIR)~\cite{rohrbach2018object} provides a systematic framework for evaluating object hallucination in image captioning tasks. CHAIR assesses caption accuracy by comparing mentioned objects against ground-truth labels, with hallucinations defined as objects present in captions but absent from ground truth. The metric operates at two levels: instance-level ($\text{CHAIR}_\text{I}$) and sentence-level ($\text{CHAIR}_\text{S}$): $\text{CHAIR}_I = \frac{\left| \{\text{hallucinated object}\}\right|}{\left| \{\text{object}\} \right|}$ and $\text{CHAIR}_S = \frac{\left| \{\text{caption w/ hallucinated objects}\} \right|}{\left| \{\text{caption}\} \right|}.$ Following established protocol~\cite{opera, pai}, we evaluate on 500 randomly sampled images from MSCOCO 2014 validation set, using the prompt ``\texttt{Please help me describe the image in detail}'' with maximum generation length of 512 tokens.

Results in Table~\ref{tab:chair} show that \ours significantly reduces hallucination in open-ended generation task, outperforming existing inference-time intervention method and other contrastive decoding methods by a substantial margin. \ours brings around 40\% relative improvement upon corresponding vanilla decoding methods. Notably, while PAI~\cite{pai} shows less efficacy for sampling-based decoding, VISTA excels across all decoding strategies. We attribute this robust performance to the contrastive design and the choice of activation space steering which is not hinged with any specific decoding strategy.

\begin{table}
    \centering
    \vspace{-10pt}
    \caption{Overall performance scores on MME full evaluation set. Higher scores indicate better general capability across perception, reasoning, and knowledge-based tasks.}
    \resizebox{0.95\linewidth}{!}{
    \begin{tabular}{llcccc}
        \toprule
        \multirow{1}{*}{\centering Decoding} & \multirow{1}{*}{\centering Method} & LLAVA-1.5 & MiniGPT-4& Shikra & InstructBLIP \\
        \midrule
        \multirow{2}{*}{Greedy} 
        & Vanilla & 1752.35 & 969.93 & 1101.50 & 1355.25 \\
        & \textbf{\ours} & \bf 1771.87 & \bf 1041.66 & \bf 1256.22 & \bf 1364.05 \\
        \midrule
        \multirow{2}{*}{Beam} 
        & Vanilla & 1749.57 & 869.74 & 1223.44 & 1357.02 \\
        & \textbf{\ours} & \bf 1763.15 & \bf 1062.48 & \bf 1323.25 & \bf 1366.57 \\
        \midrule
        \multirow{2}{*}{Nucleus} 
        & Vanilla & 1625.22 & 845.30 & 1069.60 & 1397.71 \\
        & \textbf{\ours} & \bf 1738.56 & \bf 1069.37 & \bf 1254.31 & \bf 1447.36 \\
        \bottomrule
    \end{tabular}
    }
    \label{tab:mme}
    \vspace{-15pt}
\end{table}

\textbf{POPE Evaluation.} 
The Polling-based Object Probing Evaluation (POPE)~\cite{rohrbach2018object} examines object hallucination through targeted visual questions of the form ``\texttt{Is there a <object> in the image?}''. The benchmark comprises three splits of increasing difficulty: random objects selected from a general vocabulary, frequently occurring objects chosen from common categories, and adversarially selected objects that are contextually plausible but absent from images. We evaluate on the COCO subset, reporting average accuracy and F1 scores across all splits. Since POPE evaluation is formulated as short VQA format and the response is simply \texttt{Yes} or \texttt{No}, the gradual visual information loss is not evident. We therefore adjust VSV strength to $\lambda=0.01$ to reduce VSV's impact.

Results in Table~\ref{tab:pope} demonstrate VISTA's consistent superiority across models and decoding strategies. Under greedy decoding, VISTA achieves average accuracies of 86.15\%, 77.06\%, 82.44\%, and 84.87\% for LLAVA-1.5, MiniGPT-4, Shikra, and InstructBLIP respectively, consistently outperforming both vanilla decoding and PAI. The improvements are particularly pronounced in nucleus sampling, where VISTA achieves significant gains over vanilla decoding across all models: from 81.26\% to 85.35\% (+4.09\%) for LLAVA-1.5, 60.56\% to 66.96\% (+6.40\%) for MiniGPT-4, and similarly substantial improvements for other architectures. These results highlight VISTA's particular efficacy in excavating genuine information under stochastic sampling settings, while maintaining strong performance in deterministic decoding strategies like greedy and beam search.

\begin{figure*}[t]
  \centering
   \includegraphics[width=0.9\linewidth]{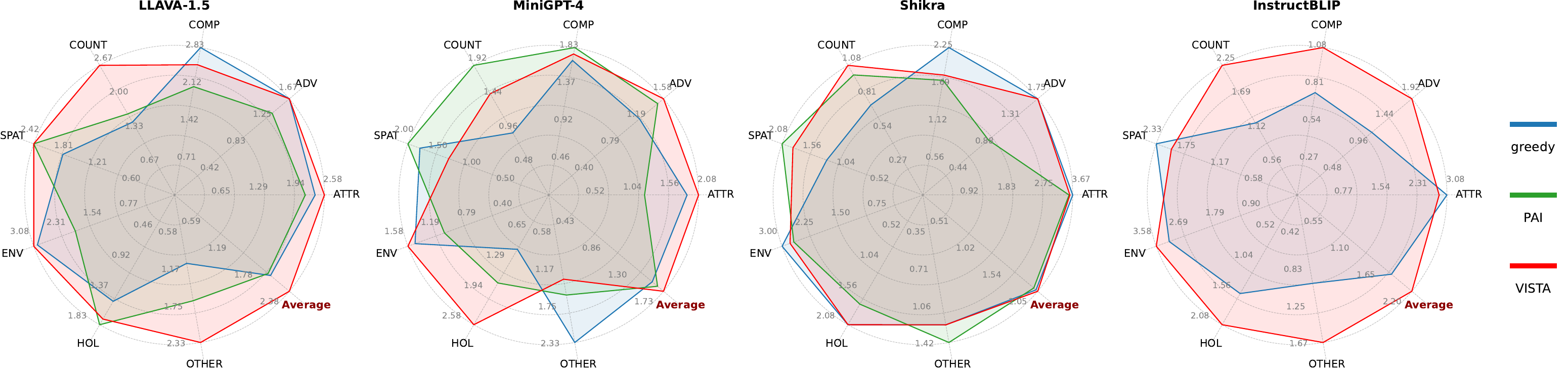}
    \vspace{-10pt}
   \caption{Performance comparison on MMHal-Bench across different question categories: attributes (ATTR), adversarial objects (ADV), comparisons (COMP), counting (COUNT), spatial relations (SPAT), environmental inference (ENV), holistic descriptions (HOL), and others (OTHER). Scores are computed using GPT-4 evaluation protocol.}
   \label{fig:mmhal}
   \vspace{-10pt}
\end{figure*}

\begin{figure}[ht]
  \centering
   \includegraphics[width=0.9\linewidth]{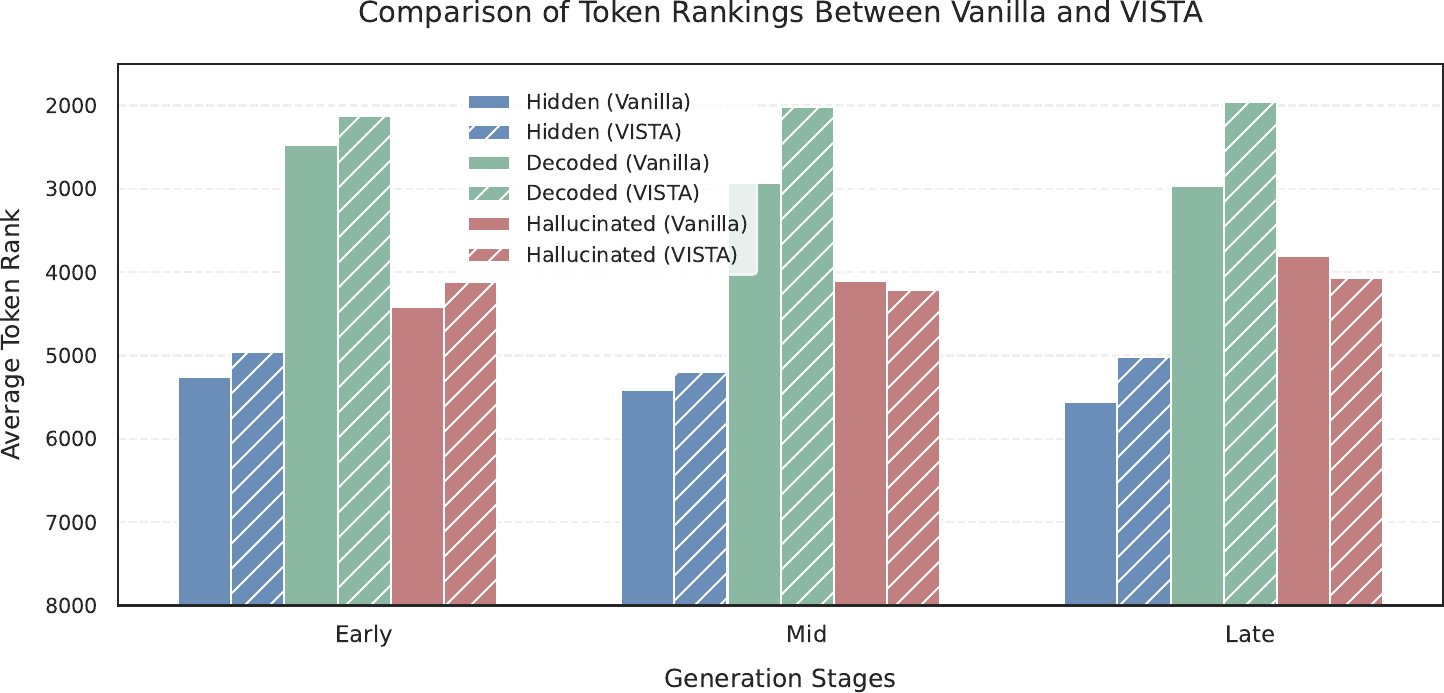}
    \vspace{-15pt}
   \caption{Cross-stage token ranking comparison between greedy and \ours on LLAVA-1.5. \ours effectively promotes the ranking of genuine tokens while depressing hallucination tokens.}
    \vspace{-15pt}
   \label{fig:compare_bar}
\end{figure}

\subsection{Results on Comprehensive Benchmarks}
\label{sec:experiment-main-comp}
We further validate VISTA on MMHal-Bench~\cite{sun2023aligning} and MME~\cite{fu2023mme}, two challenging benchmarks that examine diverse aspects of model behavior.

\textbf{MMHal-Bench Evaluation.}
MMHal-Bench~\cite{sun2023aligning} provides a specialized framework for assessing hallucination in LVLMs through 96 carefully designed image-question pairs. The benchmark spans eight distinct categories: object attributes (ATTR), adversarial objects (ADV), comparisons (COMP), counting (COUNT), spatial relations (SPAT), environmental inferences (ENV), holistic descriptions (HOL), and others (OTHER). Unlike conventional VQA evaluations, MMHal-Bench emphasizes logical reasoning and complex visual understanding, providing a rigorous test of hallucination mitigation in challenging scenarios. Model responses are evaluated using GPT-4 for alignment with ground-truth answers. We compare \ours with PAI and vanilla decoding methods for this evaluation.

The results in Fig.~\ref{fig:mmhal} demonstrate VISTA's consistent effectiveness across all evaluated LVLMs. Compared to vanilla methods, VISTA achieves substantial improvements on average scores, with LLAVA-1.5 and InstructBLIP showing the most pronounced gains ($\sim$20\% and $\sim$30\% relative improvement, respectively). VISTA particularly excels in challenging categories such as environmental inference (ENV), attribute perception (ATTR) and counting (COUNT), where visual grounding is crucial for accurate responses. While PAI shows competitive performance on specific architectures, VISTA maintains more consistent improvements across both model architectures and question types. This robust performance across diverse tasks indicates that VISTA effectively addresses hallucination while preserving general visual-language capabilities. Results under beam search and nucleus sampling (see Appendix~\ref{appendix:extra_results}) exhibit similar performance patterns, confirming VISTA's robustness across different decoding strategies.

\begin{figure*}[t]
  \centering
   \includegraphics[width=0.98\linewidth]{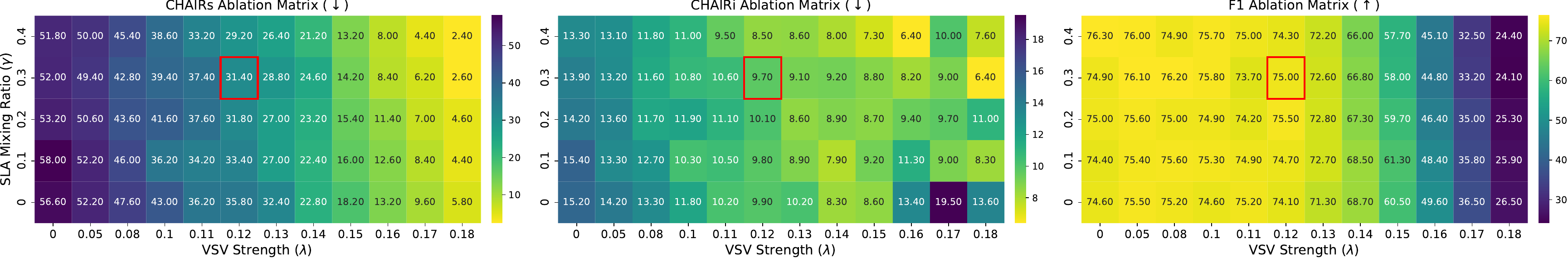}
   \vspace{-10pt}
   \caption{Ablation matrices for VSV strength ($\lambda$) and SLA mixing ratio ($\gamma$) on Shikra. Brighter color signifies the better performance. Red boxes highlight the parameter combinations we used. F1 score is included to demonstrate the overall generation quality.}
   \label{fig:ab_shikra}
   \vspace{-15pt}
\end{figure*}

\textbf{MME Evaluation.}
To assess whether hallucination mitigation affects general model capabilities, we evaluate VISTA on MME~\cite{fu2023mme}, a comprehensive benchmark encompassing 14 distinct visual-language abilities. MME tests perception, reasoning, and knowledge integration through carefully curated image-question pairs, providing a holistic view of model performance. We conduct full-set evaluation across all architectures and decoding strategies.

As shown in Table~\ref{tab:mme}, VISTA, although initially crafted for addressing hallucination, demonstrates broad benefits beyond its primary objective. Performance improvements are consistent across architectures and decoding methods, with particularly striking gains in challenging scenarios. Under nucleus sampling, MiniGPT-4's performance improves substantially from 845.30 to 1069.37, while LLAVA-1.5 maintains strong performance with significant improvement from 1625.22 to 1738.56. These comprehensive gains suggest that VISTA's approach to maintaining visual grounding enhances fundamental visual-language integration mechanisms, leading to better overall model behavior.

\subsection{On Solving Gradual Visual Information Loss}\label{sec:analysis}
Our analysis in Sec.~\ref{sec:token_analysis} identifies gradual visual information loss as a substantial challenge for long sequence generation. To validate VISTA's effectiveness in addressing this issue, we compare token logits rankings w/ and w/o \ours throughout generation. According to Fig.~\ref{fig:compare_bar}, \ours not only improves the average ranking of hidden genuine tokens throughout generation but also reverses the concerning trend of hallucinated tokens, reducing their prominence in mid and late stages where hallucination typically occurs. 
Not surprisingly, \ours also maintains a high ranking for decoded genuine tokens since VSV captures all visual clues of an image and reinforces them at all time steps. This quantitative evidence directly demonstrates VISTA's capability in maintaining visual grounding throughout the generation process. (see Appendix~\ref{appendix:extra_token_visual} for additional results).

\textbf{Case Study.} Qualitative analysis further supports our quantitative findings above. We kindly refer readers to Appendix~\ref{appendix:case_study} for a collection of qualitative examples.

\begin{table}[h]
    \centering
    \vspace{-10pt}
    \caption{Impact of window size on SLA performance. Layer ranges (X-31) indicate the span of layers used for logit augmentation, where X varies from 27 to 31. C$_\text{S}$ and C$_\text{I}$ denote CHAIR$_\text{S}$ and CHAIR$_\text{I}$ metrics,  respectively.}
    \resizebox{0.95\linewidth}{!}{
    \begin{tabular}{lccccccccccccccc}
        \toprule
        \multirow{2}{*}{\centering $\gamma$} & \multicolumn{3}{c}{31-31} & \multicolumn{3}{c}{30-31} & \multicolumn{3}{c}{29-31} & \multicolumn{3}{c}{28-31} & \multicolumn{3}{c}{27-31} \\
         
        \cmidrule(lr){2-4} \cmidrule(lr){5-7} \cmidrule(lr){8-10} \cmidrule(lr){11-13} \cmidrule(lr){14-16}
        
        & C$_\text{S}$ & C$_\text{I}$ & F1 & C$_\text{S}$ & C$_\text{I}$ & F1 & C$_\text{S}$ & C$_\text{I}$ & F1 & C$_\text{S}$ & C$_\text{I}$ & F1 & C$_\text{S}$ & C$_\text{I}$ & F1 \\
        \midrule
        0.1 & 48.2 & 13.9 & 76.2 & 48.6 & 12.7 & 77.7 & 46.8 & 12.6 & 77.4 & 46.2 & 12.2 & 74.4 & 45.8 & 11.3 & 77.6\\
        0.2 & 56.6 & 16.4 & 75.3 & 49.4 & 14.4 & 76.5 & 47.4 & 12.7 & 77.3 & 46.8 & 12.1 & 77.7 & 43.2 & 11.7 & 77.6\\
        0.3 & 62.0 & 18.8 & 72.9 & 55.4 & 15.7 & 75.9 & 49.2 & 14.2 & 76.5 & 45.8 & 12.4 & 77.9 & \cellcolor{lightgray} \bf 42.8 & \cellcolor{lightgray} \bf 11.3 & \cellcolor{lightgray} \bf78.4 \\
        0.4 & 61.2 & 18.2 & 73.3 & 57.6 & 15.7 & 75.3 & 52.6 & 14.5 & 76.1 & 48.8 & 13.5 & 77.0 & 46.6 & 12.3 & 77.2 \\
        \bottomrule
    \end{tabular}
    }
    \label{tab:ab_window_size}
    \vspace{-15pt}
\end{table}

\subsection{Ablation Study}
\label{sec:ablation}
To thoroughly investigate the effectiveness of \ours, we gauge the practical latency of \ours, and analyze how different VSV strength ($\lambda$) and SLA mixing ratio ($\gamma$) affect the model's performance in terms of hallucination reduction (CHAIR-S and CHAIR-I metrics) and overall quality (F1 score). Results are in Fig.~\ref{fig:ab_shikra}. Additional ablation results are deferred to Appendix~\ref{appendix:extra_ablation}.

\begin{table}[h]
    \centering
    \vspace{-10pt}
    \caption{Measure of throughput and latency on LLAVA-1.5. Greedy decoding strategy is applied and listed as baseline.}
    \resizebox{0.98\linewidth}{!}{
    \begin{tabular}{l|c|ccc} 
        \toprule
        Methods & Greedy & VCD & PAI &  \textbf{\ours} (ours) \\
        \midrule
        Latency (ms/token) $\downarrow$ & 28.54 ($\times1.0$) & 58.34 ($\times2.04$) & 57.78 ($\times2.02$) & \bf 36.32 ($\times1.27$) \\
        Throughput (token/s) $\uparrow$ & 35.04 ($\times1.0$) & 17.14 ($\times0.49$) & 17.31 ($\times 0.49$) & \bf 27.53 ($\times 0.79$) \\
        \bottomrule
    \end{tabular}
    }
    \vspace{-5pt}
    \label{tab:latency}
\end{table}

\textbf{Efficiency.} The greedy decoding latency in Table~\ref{tab:latency} shows that \ours is more efficient than the other test-time intervention strategy PAI and contrastive decoding strategy like VCD.

\textbf{VSV Strength ($\lambda$)}. As we vary injection strength $\lambda$ from 0.0 to 0.18, both CHAIR-S and CHAIR-I scores improve significantly. However, using inappropriate scale (see Fig.~\ref{fig:ab_shikra}) will cause clear degradation in F1, suggesting overemphasis on visual features can harm generation quality. 

\textbf{SLA Mixing Ratio ($\gamma$)}. The impact of $\gamma$ is studied across values from 0 to 0.4, revealing that moderate values of $\gamma$ (0.2-0.3) yield the best balance between hallucination reduction and generation quality, and higher $\gamma$ values ($\leq$ 0.4) leads to degraded performance, hinting on the importance of syntactic information.

\textbf{Synergy \& Robustness.} 
As visualized in Fig.~\ref{fig:ab_shikra}, there exists a synergistic relation between $\lambda$ and $\gamma$. Moderate values of both parameters consistently outperform extreme settings of either component alone. Notably, generation quality remains stable (F1$>$72.0) across a broad range of parameter combinations, demonstrating the robustness of our approach. This complementarity suggests that VSV and SLA address different aspects of the hallucination problem.

\textbf{Window Size in SLA ($w$)}. In Table~\ref{tab:ab_window_size}, we explore window size from one layer up to five layers across different mixing ratios. As shown, larger window size generally reduces hallucination, and the optimal configuration achieves both strong hallucination reduction and the highest F1 score (78.4). Interestingly, we observe an inverse relationship between window size and optimal $\gamma$ values, suggesting that while broader layer spans capture richer visual information, they require more conservative mixing ratios to maintain generation stability.

\section{Related Work}
\label{sec:related_work}

\textbf{Hallucination Mitigation in LVLMs.} 
Hallucination -- the generation of content that is irrelevant, factually incorrect, or inconsistent with visual inputs~\cite{bai2024hallucination} -- represents a fundamental challenge in LVLM development. Research has identified three primary sources: limitations in visual encoder capabilities~\cite{tong2024eyes, liu2024llavanext, shi2024eagle}, excessive reliance on learned parametric knowledge~\cite{li-etal-2023-evaluating, zhou2023analyzing, vcd, opera}, and noisy training data~\cite{liu2023mitigating, yu2024hallucidoctor}. Mitigation approaches span training-based solutions with refined datasets~\cite{yue2024less, jiang2024hallucination}, post-processing techniques including revision~\cite{yin2023woodpecker, zhou2023analyzing} and verification~\cite{chen2024halc, sun2023aligning}, and inference-time interventions like Visual Contrastive Decoding~\cite{vcd} and enhanced attention methods~\cite{pai}. Recent studies revealing ``text inertia''~\cite{pai}, where models generate similar hallucinations without visual input, highlight concerning reliance on learned text patterns. Similar to VISTA, VTI~\cite{liu2024reducing} also applies representation engineering technique to LVLM hallucination domain, yet with a focus on the instability issue of visual representations. While prior findings advance our understanding, how hallucination propagates through model architectures remains elusive, and existing solutions often require external supervision or are hinged with specific decoding strategies.

\textbf{Contrastive Decoding in LVLMs.} 
Contrastive decoding, originally introduced in NLP~\cite{li2022contrastive, shi2023trusting}, has emerged as a promising approach for reducing hallucination in LVLMs. Recent adaptations of this technique have explored various contrasting strategies: VCD~\cite{vcd} introduces visual-specific contrasts by crafting noisy visual tokens as negative samples, while DoLa~\cite{dola} innovates by contrasting logits distributions from different layers within the same model, using divergence measurements to dynamically select contrasting layers. Taking a temporal perspective, M3ID~\cite{favero2024multi} proposes a "horizontal" strategy that contrasts current logits with those from previous timesteps. Other approaches extend contrastive techniques to attention mechanisms~\cite{woo2024don}. While these methods primarily operate in the logits space, our \ours takes a different approach by performing contrasts in the activation space and intervening at residual streams. This earlier-stage intervention strategy offers an efficient alternative that can complement existing decoding methods.

\section{Conclusion and Limitations}\label{sec:conclusion}
This study investigates the hidden life of tokens in Large Vision Language Models (LVLMs) and introduces \textbf{\ours}~(\textbf{V}isual \textbf{I}nformation \textbf{S}teering with \textbf{T}oken-logit \textbf{A}ugmentation), a lightweight approach to mitigate hallucination.
Through systematic analysis, we reveal that visual information gradually attenuates during text generation, but can be effectively restored through our framework's visual information steering and strategic use of early-layer logits. Extensive experimentation across diverse architectures and decoding strategies demonstrates that our framework significantly reduces hallucination while preserving generation quality.
These findings not only illuminate the hidden dynamics of LVLM behavior but also establish visual information steering as a promising direction for enhancing the reliability of multimodal AI systems.

\textbf{Limitations.} VISTA is subject to several limitations. First, while VISTA demonstrates robustness across a range of hyperparameter values, optimal settings may vary across different architectures. Second, the effectiveness of VSV relies on the quality of visual cues extracted by the LVLM's vision encoder -- models with weak visual encoding capabilities may see reduced benefits. Third, the current implementation focuses on addressing hallucination in single-round tasks; adaptation to interactive scenarios like visual dialogue may require additional considerations.

\section*{Impact Statement}
This research advances methods for making large vision-language models more trustworthy and reliable through mitigating hallucination. While the proposed method demonstrates promising results, its effectiveness is subject to the inherent capability of large vision-language model, and improper usage may adversely affect model's performance. To the best of our knowledge, there are no ethical or other concerns that need to be addressed.


\bibliography{example_paper}
\bibliographystyle{icml2025}

\newpage
\appendix
\onecolumn

\begin{table*}[ht]
\footnotesize
\centering
\begin{tabular}{p{160mm}}
    \toprule
    GPT-4o Prompt
    \\
    \midrule
    You are a vision-language evaluator. Given an image and an AI-generated description, perform the following tasks:\\
    \\
    1. List clearly visible contents in the image that are not mentioned in the description.\\
    2. List hallucinated contents in the description that are not present in the image.\\
    3. List contents accurately described in the description that match the image.\\
    \\
    For each task, include objects, object properties (e.g., color, count, position), and relationships between objects. You must answer each content with a single word, separating different contents by commas. If no contents apply, write "None". Make sure there is no overlapping words between three tasks.\\
    \\
    
    Answer 1: [Missing contents]\\
    Answer 2: [Hallucinated contents]\\
    Answer 3: [Accurate contents]\\
    \bottomrule
\end{tabular}
\caption{The prompt used for GPT-4o to identify genuine and hallucinated words.}
\label{tab:gpt4o_prompt}
\end{table*}


\begin{algorithm}[t]
\caption{Token Ranking Analysis Framework}
\label{alg:token_rank}
\begin{algorithmic}[1]
\REQUIRE Image $I$, LVLM model $M$, Oracle model $O$
\ENSURE Token ranking matrices $R$ for each category
\STATE /* Generate description using LVLM */
\STATE $D \leftarrow M(I)$ 

\STATE /* Classify tokens using Oracle */
\STATE $W_{dec}, W_{hid}, W_{hal} \leftarrow O(I, D)$ \COMMENT{decoded, hidden, hallucinated}
\STATE $T_{dec}, T_{hid}, T_{hal} \leftarrow \text{Tokenize}(W_{dec}, W_{hid}, W_{hal})$

\FOR{each token category $T_c$ in $\{T_{dec}, T_{hid}, T_{hal}\}$}
    \FOR{$t = 1$ to $|D|$}
        \FOR{$l = 1$ to $L$}
            \STATE $h_t^l \leftarrow \text{GetHiddenState}(M, l, t)$ \COMMENT{Eq. 3}
            \STATE $\text{logits} \leftarrow H(h_t^l)$ \COMMENT{Eq. 4}
            \FOR{each token $x$ in $T_c$}
                \STATE $R_t^l(x) \leftarrow \text{rank}(\text{logits}, x)$ \COMMENT{per Eq. 4}
            \ENDFOR
        \ENDFOR
    \ENDFOR
\ENDFOR

\STATE /* Compute stage-wise aggregation */
\FOR{each stage $s$ in $\{\text{early}, \text{mid}, \text{late}\}$}
    \FOR{each category $c$ in $\{\text{dec}, \text{hid}, \text{hal}\}$}
        \STATE $T_s \leftarrow \text{GetTimeSteps}(s)$ \COMMENT{timesteps in stage $s$}
        \STATE $L_f \leftarrow \text{GetFinalLayers}()$ \COMMENT{final 5 layers}
        \STATE $\bar{R}_s^c \leftarrow \text{mean}(\{R_t^l(x) \mid t \in T_s, l \in L_f, x \in T_c\})$
    \ENDFOR
\ENDFOR

\OUTPUT $R$ \COMMENT{Ranking matrices for all categories}
\end{algorithmic}
\end{algorithm}

\section{Implementation Details for Token Ranking Analysis}\label{appendix:token_analysis}

\subsection{Prompt for GPT-4o}
We employ GPT-4o as our oracle model for identifying three categories of tokens: hidden genuine tokens, decoded genuine tokens, and hallucinated tokens. The precise prompting strategy used to elicit these classifications is detailed in Table~\ref{tab:gpt4o_prompt}.

\subsection{Additional Implementation Details}
A detailed token analysis algorithm is provided in Algorithm~\ref{alg:token_rank}.

\noindent{\textbf{Token Processing.}} As shown in Algorithm~\ref{alg:token_rank}, we derive genuine and hallucinated tokens from their corresponding word-level classifications. In cases where a single word decomposes into multiple tokens under the model's tokenization scheme, we adopt the first token as a representative proxy for the entire word. This approach ensures consistent handling of multi-token words while maintaining analytical tractability.

\noindent{\textbf{Ranking Aggregation Protocol.}} When computing cross-stage token rankings (visualized in Fig.~\ref{fig:global_view} left), we implement a focused aggregation strategy that considers only the final five layers of the model, deliberately excluding rankings from earlier layers. This methodological choice mitigates the inherent embedding disparity between the model's decoding layer and preceding layers. Since the LLM's decoding head is specifically trained on final-layer hidden states, the reliability of token rankings decreases with distance from this layer. Consequently, we restrict our analysis to a window of layers proximate to the final layer to ensure robust and meaningful ranking estimates.

\subsection{Advantages of Token Ranking Analysis}\label{appendix:token_analysis_discussion}

\noindent\textbf{Why Not Attention?} While attention matrices have been extensively studied in LVLM hallucination research to understand information flow patterns and inform mitigation strategies~\cite{opera, pai}, our token ranking methodology offers several distinct advantages. Previous work, such as PAI~\cite{pai}, has attributed hallucination phenomena like ``text inertia'' to insufficient attention allocation to visual tokens. However, this interpretation is potentially confounded by the presence of ``anchor tokens''~\cite{wang-etal-2023-label, opera} that aggregate and redistribute information across the network. The existence of these information hubs means that reduced attention weights on visual tokens, particularly in later layers where visual information can be accessed indirectly through anchor positions, may not necessarily indicate information loss.

Token ranking analysis, by contrast, provides more direct insights into the model's processing of visual information. Through explicit tracking at the token level, this approach enables quantitative measurement of visual information preservation throughout the generation process. The methodology reveals gradual visual information degradation patterns that might be obscured in attention-based analyses. Furthermore, token ranking analysis uncovers previously unobserved phenomena such as hidden genuine information and early excitation patterns, which are not readily distinguishable through attention analysis alone. These capabilities make token ranking analysis particularly well-suited for investigating the mechanisms underlying hallucination in LVLMs.

\noindent\textbf{Limitations.} Our token ranking analysis approach also pose certain limitations:

\begin{enumerate}
    \item \textbf{Oracle Model Reliability:} While GPT-4o serves as our oracle for identifying genuine and hallucinated content, this process can introduce potential biases and uncertainties. The oracle model's classifications may not perfectly align with human judgments, and its own biases could influence the categorization of tokens. This is particularly challenging for nuanced cases where the distinction between genuine and hallucinated content is subtle.
    
    \item \textbf{Embedding Space Discrepancy:} Another limitation arises from applying the LLM's decoding head to hidden states from earlier layers. Since the decoding head is specifically trained on final-layer representations, there exists an embedding space misalignment when analyzing preceding layers. This discrepancy becomes more pronounced for layers distant from the final layer, potentially leading to less reliable token rankings in earlier stages of the network. While our analysis mitigates this by focusing on layers proximate to the last layer, the issue remains inherent to the methodology.
\end{enumerate}

Despite above limitations, our analysis provides valuable insights into LVLM's behavior and has proven effective in motivating our hallucination mitigation approach.

\subsection{Additional Token Ranking Analysis}\label{appendix:extra_token_visual}

\begin{figure}
  \centering
   \includegraphics[width=0.5\linewidth]{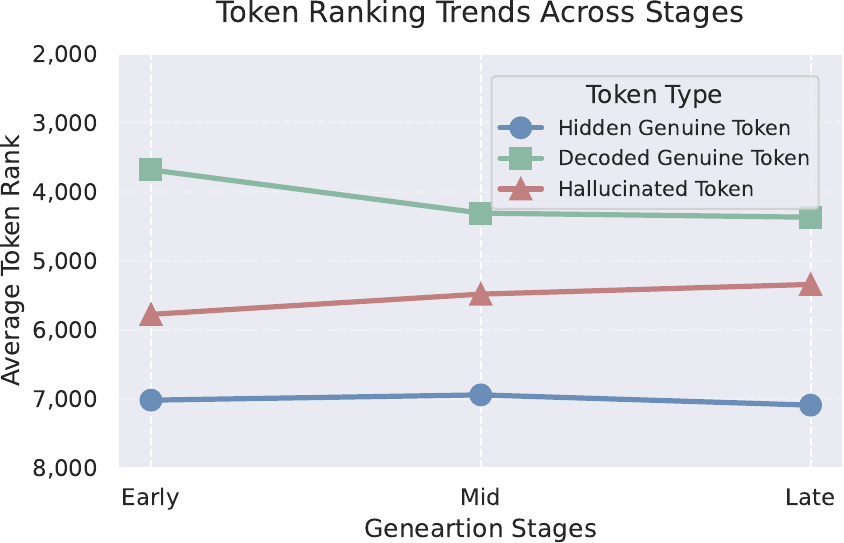}
   \caption{Cross-stage token ranking on Shikra.}
   \label{fig:shikra_rank_trends}
\end{figure}

\begin{figure}
  \centering
   \includegraphics[width=0.5\linewidth]{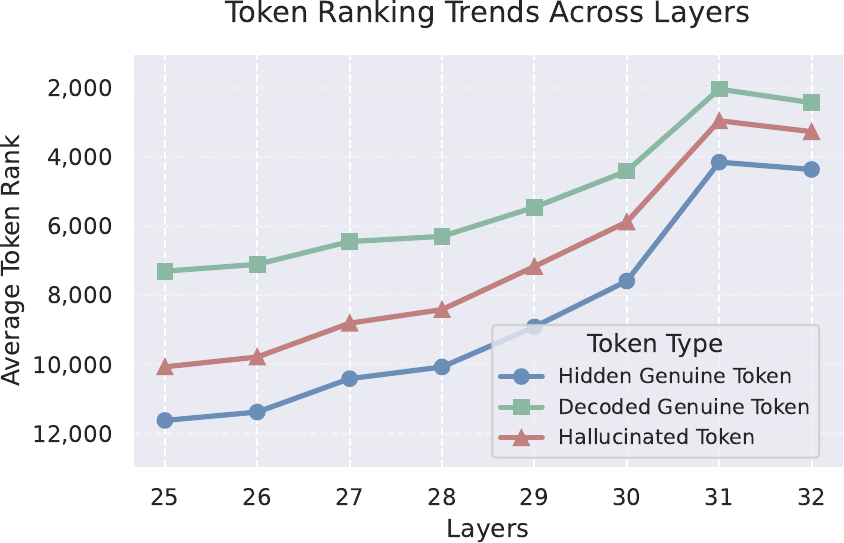}
   \caption{Layer-wise token rankings on Shikra.}
   \label{fig:shikra_early_excitation}
\end{figure}

\begin{figure}
  \centering
   \includegraphics[width=0.5\linewidth]{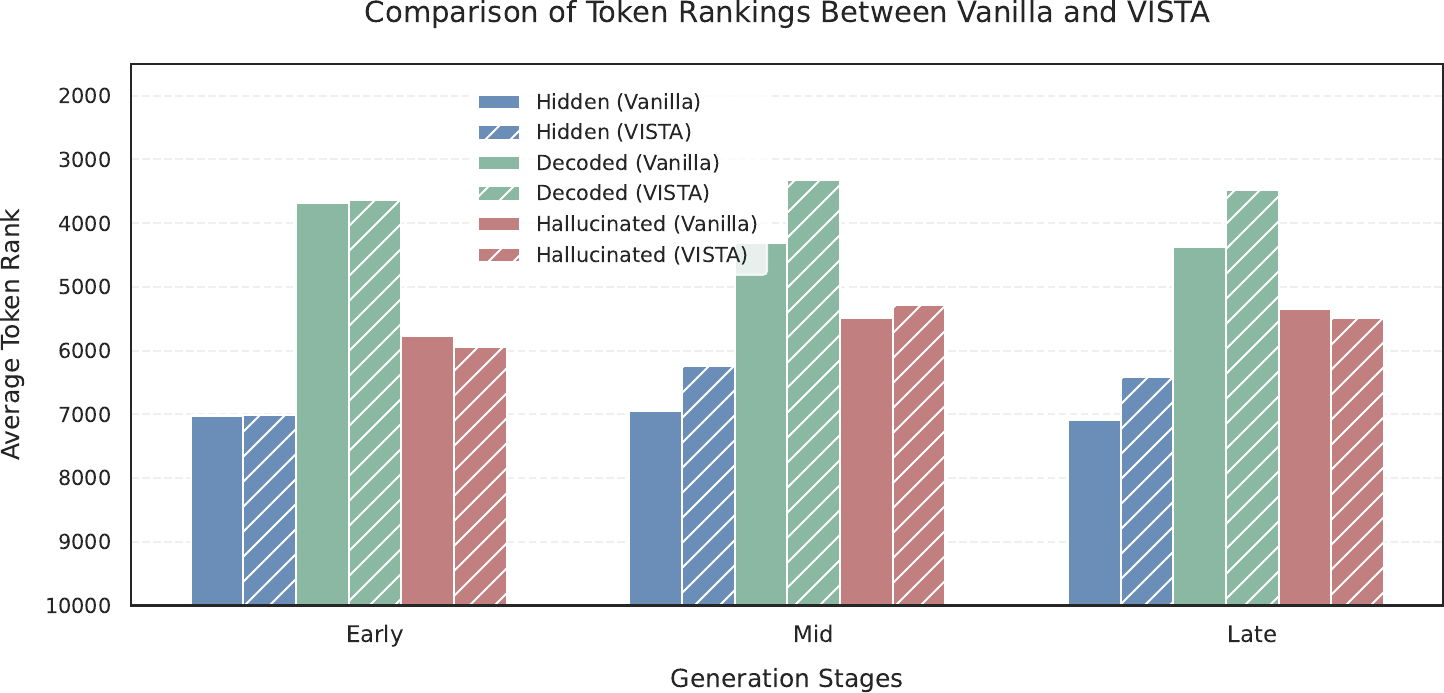}
   \caption{Cross-stage token ranking comparison between greedy and \ours (greedy-based) on Shikra. \ours effectively promotes the ranking of genuine tokens while depressing hallucination tokens.}
   \label{fig:shikra_compare_bar}
\end{figure}

In the main text, we present token ranking analysis results for LLAVA-1.5. Here, we extend this analysis to other architectures to demonstrate the generalizability of our observations. As shown in Fig.~\ref{fig:shikra_rank_trends}, the cross-stage token ranking analysis on Shikra exhibits similar patterns to those observed in LLAVA-1.5, with genuine tokens experiencing gradual rank degradation while hallucinated tokens become increasingly prioritized across generation stages. The layer-wise analysis presented in Fig.~\ref{fig:shikra_early_excitation} further corroborates the early excitation phenomenon, where semantic tokens achieve peak activation in layers preceding the final decoding layer. Fig.~\ref{fig:shikra_compare_bar} presents a comparative analysis between vanilla decoding and VISTA (greedy-based) on Shikra. The results demonstrate VISTA's effectiveness in maintaining the ranking of genuine tokens throughout the generation process while simultaneously suppressing the promotion of hallucinated tokens. This pattern is consistent with our findings for LLAVA-1.5, suggesting that the phenomena we identified and the effectiveness of our mitigation strategy generalize across different LVLM architectures. The consistency of these patterns across architectures with distinct design choices (linear projector in Shikra versus Q-former in other models) provides strong evidence for the fundamental nature of these phenomena in LVLM generation dynamics.

\section{Additional Experiments}\label{appendix:extra_results}

\subsection{MMHal-Bench Results For Other Decoding Strategies}
\begin{figure*}
  \centering
   \includegraphics[width=1.0\linewidth]{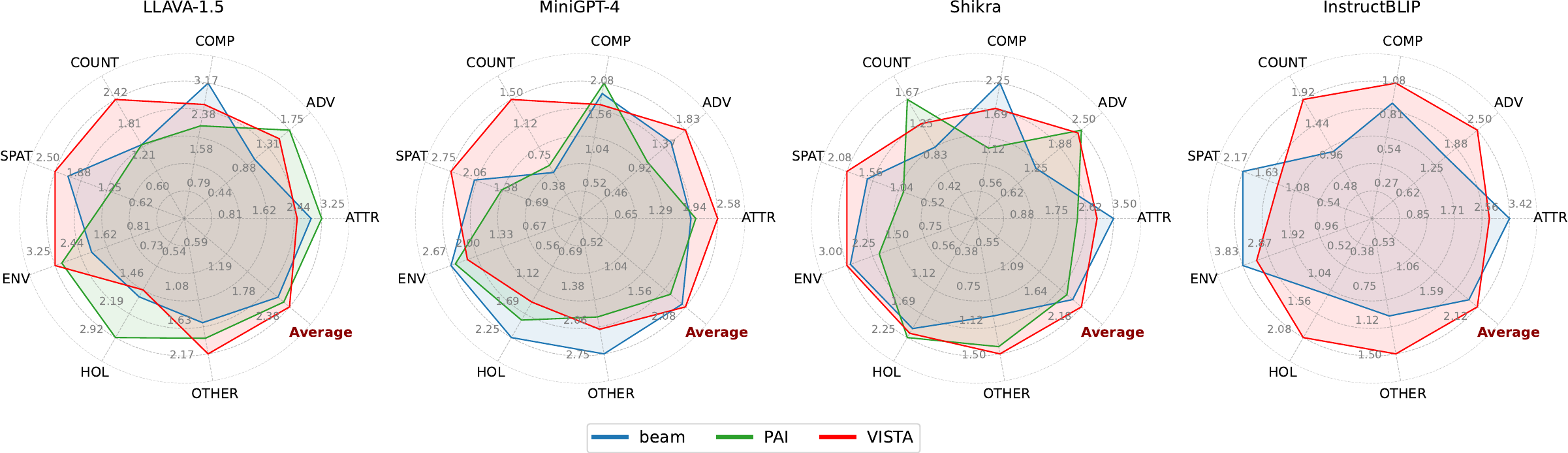}
   \caption{Performance comparison on MMHal-Bench using beam search.}
   \label{fig:beam_mmhal}
\end{figure*}

\begin{figure*}
  \centering
   \includegraphics[width=1.0\linewidth]{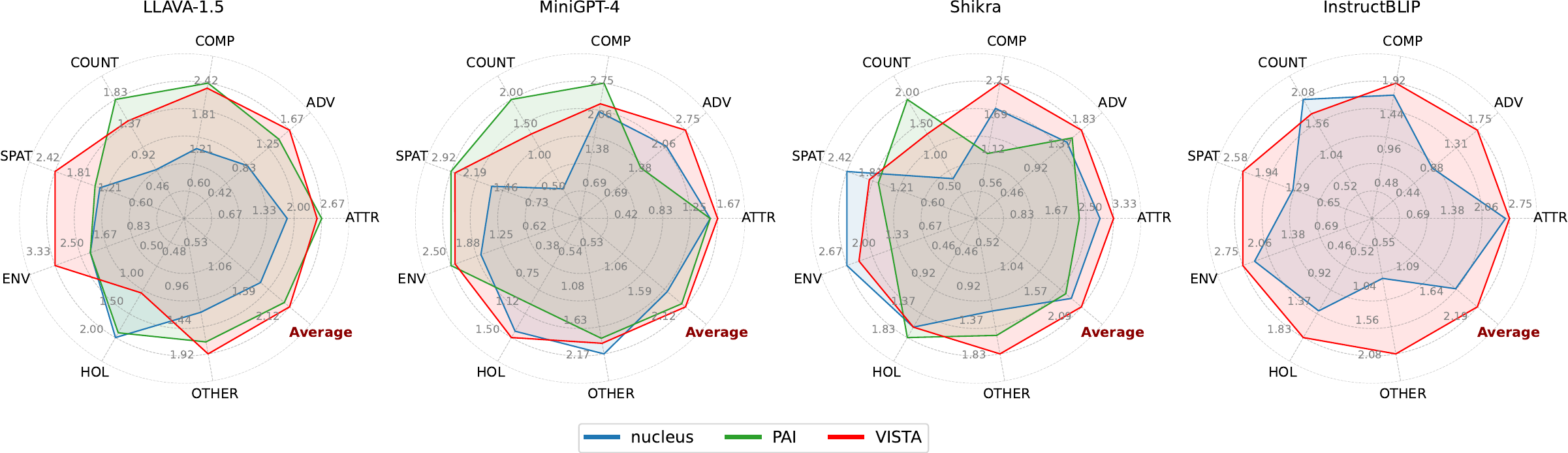}
   \caption{Performance comparison on MMHal-Bench using nucleus sampling.}
   \label{fig:nucleus_mmhal}
\end{figure*}

We further report results of MMHal-Bench under beam search (Fig.~\ref{fig:beam_mmhal}) and nucleus sampling (Fig.~\ref{fig:nucleus_mmhal}). As demonstrated in figures, VISTA consistently improves overall performance across all evaluated LVLMs under both decoding strategies. The performance trends remain consistent with those observed under greedy decoding in the main text, further validating the robustness of our approach across different inference strategies. These comprehensive results demonstrate that VISTA's mechanisms for maintaining visual grounding and promoting semantic richness are effective regardless of the chosen decoding strategy.

\begin{figure}
    \centering
    \includegraphics[width=1.0\linewidth]{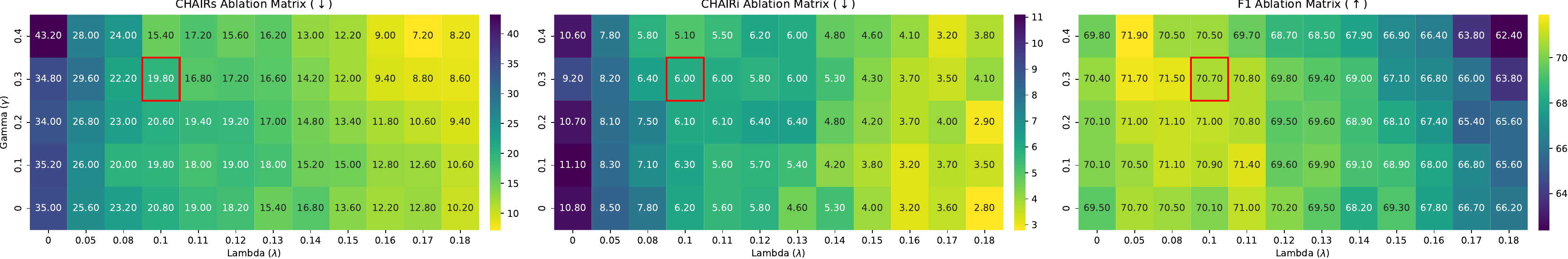}
    \caption{Ablation matrices for VSV strength ($\lambda$) and SLA mixing ratio ($\gamma$) on MiniGPT-4. Brighter color signifies the better performance, and red boxes highlight the parameter combinations used in Table~\ref{tab:chair}. F1 score is included to indicate the overall generation quality.}
    \label{fig:ab_minigpt4_greedy}
\end{figure}

\begin{figure}
    \centering
    \includegraphics[width=1.0\linewidth]{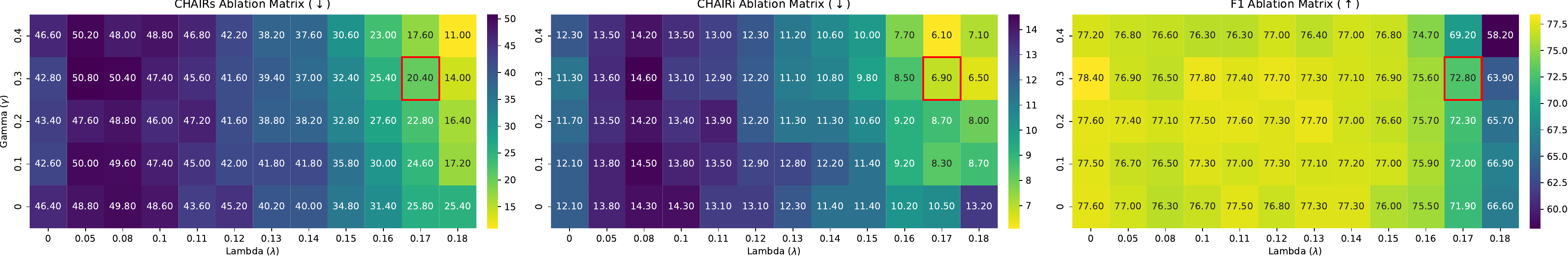} 
    \caption{Ablation matrices for VSV strength ($\lambda$) and SLA mixing ratio ($\gamma$) on LLAVA-1.5. Brighter color signifies the better performance, and red boxes highlight the parameter combinations used in Table~\ref{tab:chair}. F1 score is included to indicate the overall generation quality.}
    \label{fig:ab_llava_greedy}
\end{figure}

\begin{figure}
    \centering
    \includegraphics[width=1.0\linewidth]{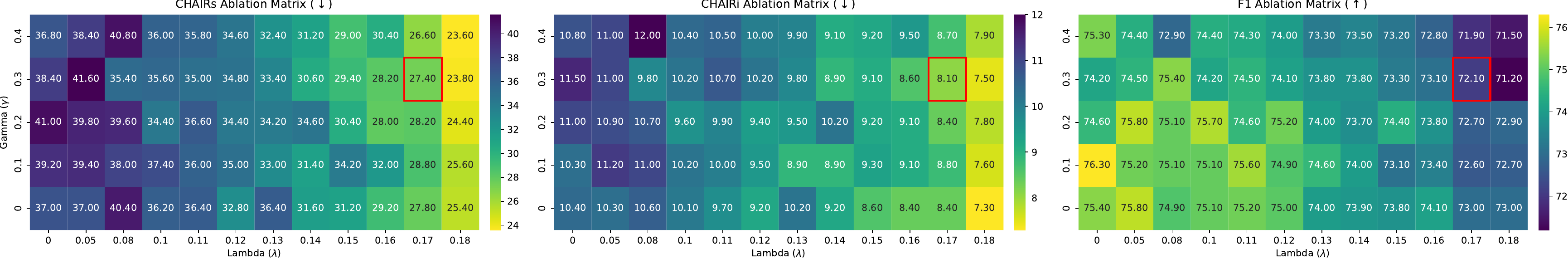} 
    \caption{Ablation matrices for VSV strength ($\lambda$) and SLA mixing ratio ($\gamma$) on InstructBLIP. Brighter color signifies the better performance, and red boxes highlight the parameter combinations used in Table~\ref{tab:chair}. F1 score is included to indicate the overall generation quality.}
    \label{fig:ab_instructblip_greedy}
\end{figure}

\subsection{Additional Ablation Results}\label{appendix:extra_ablation}

In addition to the ablation studies presented in the main text, we further provide detailed ablation results on LLAVA-1.5, MiniGPT-4, and InstructBLIP to validate our hyperparameter choices. Figure~\ref{fig:ab_minigpt4_greedy} presents the ablation matrices for MiniGPT-4, examining the impact of VSV strength ($\lambda$) and SLA mixing ratio ($\gamma$) on CHAIR$_\text{S}$, CHAIR$_\text{I}$, and F1 scores. The results reveal similar trends to those observed in Shikra, though with a slightly lower optimal $\lambda$ value of 0.1, suggesting architecture-specific sensitivity to visual steering. For LLAVA-1.5 (Figure~\ref{fig:ab_llava_greedy}), the ablation matrices demonstrate particularly strong performance improvements with stronger VSV value ($\lambda = 0.17$). The InstructBLIP results (Figure~\ref{fig:ab_instructblip_greedy}) show robust performance across a broader range of parameter combinations, with optimal performance achieved at $\lambda = 0.17$ and $\gamma = 0.3$, matching the configuration used for LLAVA-1.5.

Across all architectures, we observe a consistent pattern where moderate values of both VSV strength and SLA mixing ratio yield the best balance between hallucination reduction and generation quality. This consistency, despite architectural differences, validates the generality of our approach while highlighting the need for architecture-specific fine-tuning of hyperparameters for optimal performance.

\section{Case Study}\label{appendix:case_study}
\begin{figure*}[t]
    \centering
    \includegraphics[width=0.95\textwidth]{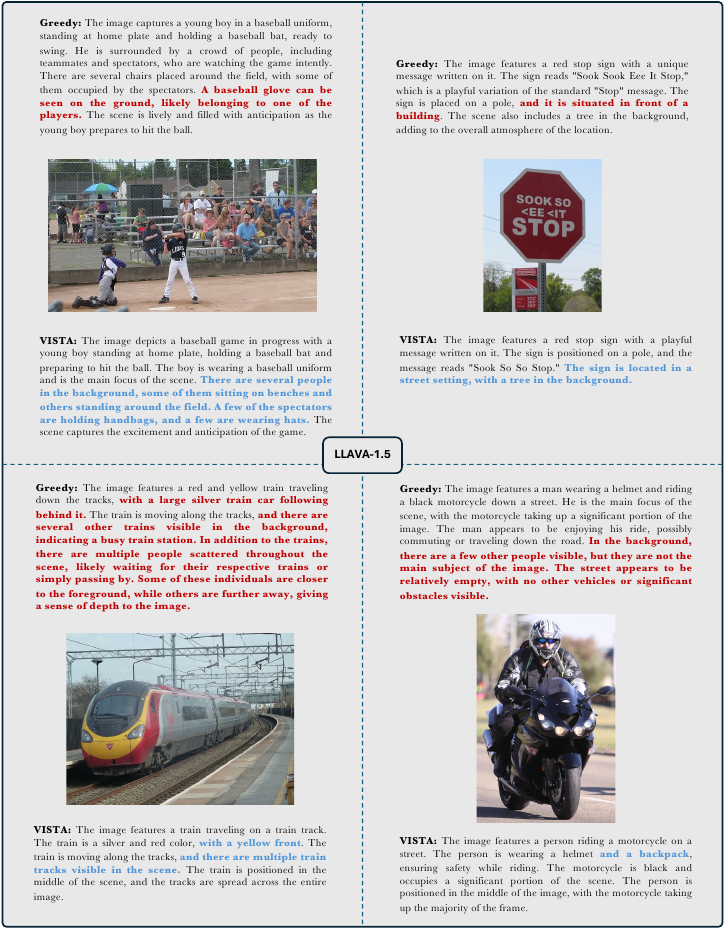} 
    \caption{Case study for LLAVA-1.5.}
    \label{fig:cs_llava_greedy}
\end{figure*}

\begin{figure*}[t]
    \centering
    \includegraphics[width=0.95\textwidth]{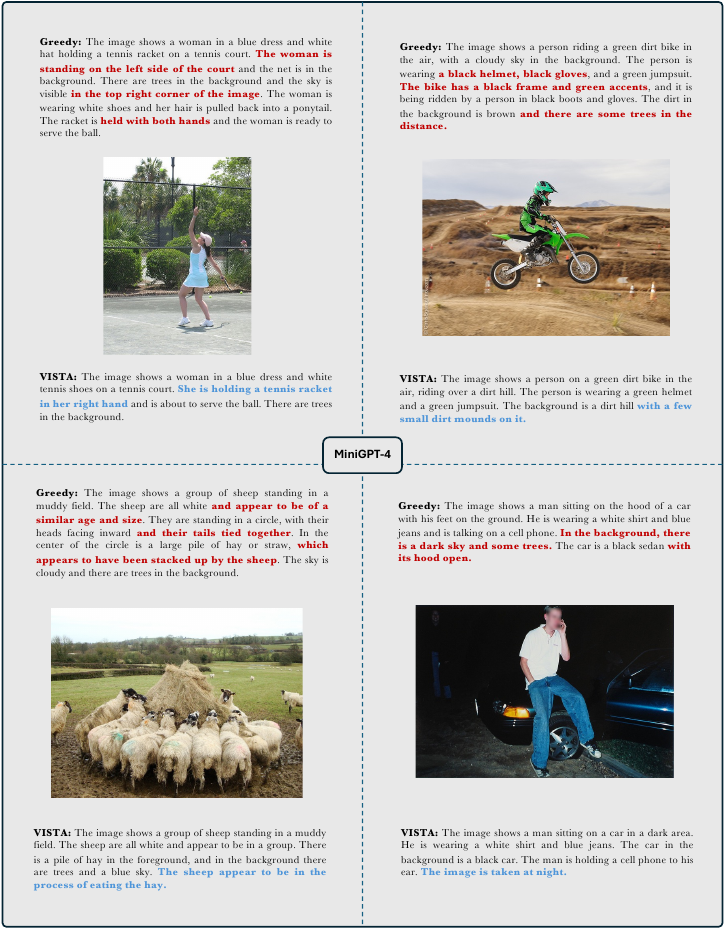} 
    \caption{Case study for MiniGPT-4.}
    \label{fig:cs_minigpt4_greedy}
\end{figure*}

\begin{figure*}[t]
    \centering
    \includegraphics[width=0.95\textwidth]{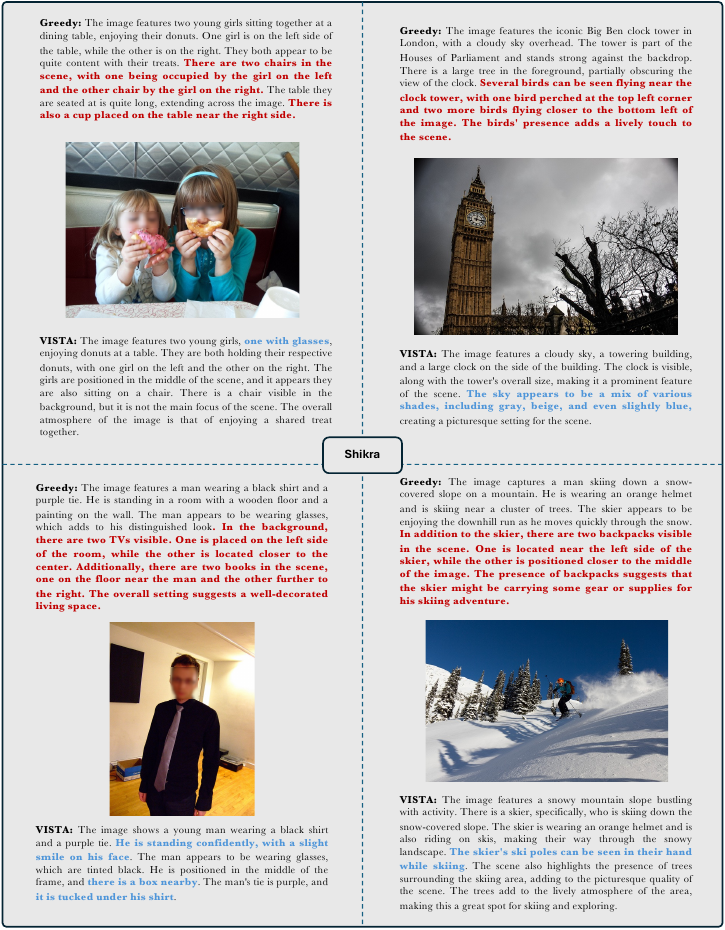} 
    \caption{Case study for Shikra.}
    \label{fig:cs_shikra_greedy}
\end{figure*}

\begin{figure*}[t]
    \centering
    \includegraphics[width=0.95\textwidth]{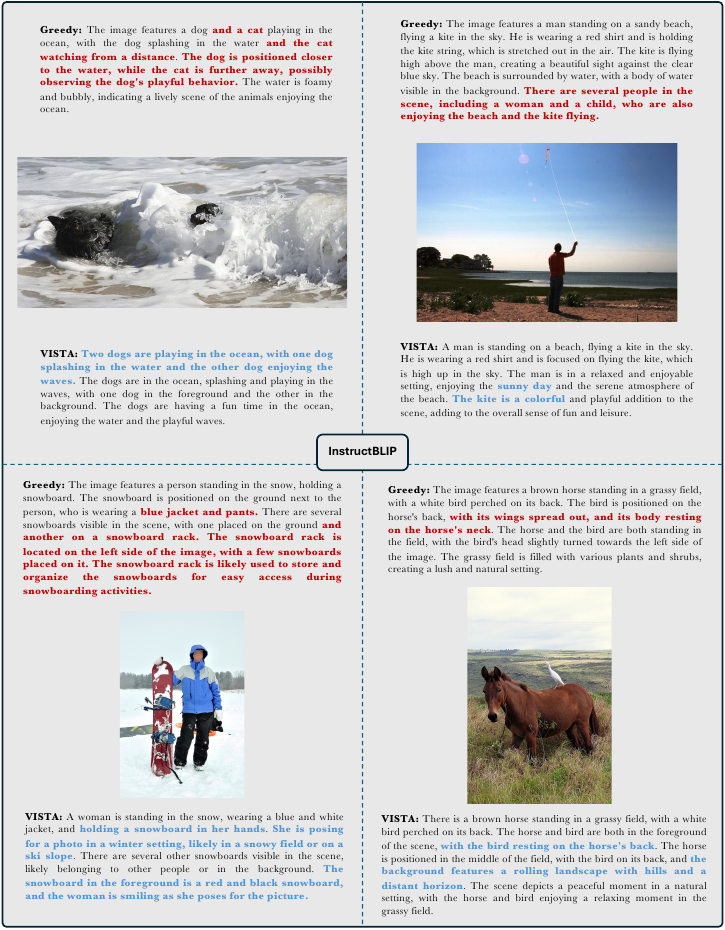} 
    \caption{Case study for InstructBLIP.}
    \label{fig:cs_instructblip_greedy}
\end{figure*}

In this section, we extend our qualitative evaluation across all four architectures to demonstrate VISTA's effectiveness in reducing hallucination while promoting genuine information. Figures~\ref{fig:cs_llava_greedy}-\ref{fig:cs_instructblip_greedy} present comparative examples between vanilla decoding and VISTA for LLAVA-1.5, MiniGPT-4, Shikra, and InstructBLIP respectively. For each example, hallucinated content is highlighted in \textcolor[HTML]{C00400}{\textbf{red}}, while genuine information that was previously omitted or wrongly described but successfully recovered by VISTA is marked in \textcolor[HTML]{4E95D9}{\textbf{blue}}. The examples demonstrate VISTA's consistent ability to reduce hallucination across different architectures while eliciting pertinent visual details that vanilla decoding fails to capture.

\label{sec:cs}

\end{document}